\documentclass[letterpaper, 10 pt, conference]{ieeeconf}  

\IEEEoverridecommandlockouts                              

\usepackage{graphics} 

\usepackage{algorithm}
\usepackage{algpseudocode}
\usepackage{amsmath} 
\usepackage{amssymb}
\usepackage{url}
\usepackage{arydshln}
\usepackage{newfloat}
\usepackage{listings}
\usepackage{siunitx}
\usepackage{xspace}
\usepackage{multicol}
\usepackage{booktabs}      
\usepackage{multirow}      
\usepackage[table]{xcolor} 
\usepackage{graphicx}      
\graphicspath{ {image/} }  
\usepackage{xcolor}
\usepackage{makecell} 
\title{\LARGE \bf
    QuadFM: Foundational Text-Driven Quadruped Motion Dataset for Generation and Control
}

\author{Li Gao$^{\ast}$, Fuzhi Yang$^{\ast}$, Jianhui Chen$^{\ast}$, Liu Liu$^{\dag}$, Yao Zheng, Yang Cai, Ziqiao Li \\
\emph{\{liangliang.gl, fuzhi.yfz, yunyuan.cjh, diana.ll, yc498306\}@alibaba-inc.com} \\
\emph{yangcai.cy@autonavi.com} \\
\thanks{$^{\ast}$Equal Contribution}
\thanks{$^{\dag}$Corresponding Author}
\thanks{AMAP, Alibaba Group}
}

\begin{document}

\maketitle
\thispagestyle{empty}
\pagestyle{empty}

\begin{abstract}

Despite significant advances in quadrupedal robotics, a critical gap persists in foundational motion resources that holistically integrate diverse locomotion, emotionally expressive behaviors, and rich language semantics—essential for agile, intuitive human-robot interaction. Current quadruped motion datasets are limited to a few mocap primitives (e.g., walk, trot, sit) and lack diverse behaviors with rich language grounding. To bridge this gap, we introduce Quadruped Foundational Motion (QuadFM) , the first large-scale, ultra-high-fidelity dataset designed for text-to-motion generation and general motion control. QuadFM contains 11,784 curated motion clips spanning locomotion, interactive, and emotion-expressive behaviors (e.g., dancing, stretching, peeing), each with three-layer annotation—fine-grained action labels, interaction scenarios, and natural language commands—totaling 35,352 descriptions to support language-conditioned understanding and command execution.

We further propose Gen2Control RL, a unified framework that jointly trains a general motion controller and a text-to-motion generator, enabling efficient end-to-end inference on edge hardware. On a real quadruped robot with an NVIDIA Orin, our system achieves real-time motion synthesis ($<$500 ms latency). Simulation and real-world results show realistic, diverse motions while maintaining robust physical interaction. The dataset will be released at \url{https://github.com/GaoLii/QuadFM}.
\end{abstract}

\section{INTRODUCTION}
Quadrupedal robots, as bio-inspired legged systems, have become a key platform for embodied intelligence and robot locomotion research due to their superior stability, agility, and adaptability to unstructured terrain~\cite{margolis2024rapid}. The long-term goal of this field goes beyond achieving high-performance steady walking and obstacle negotiation; it is to endow robots with animal-like behaviors that are natural, temporally coherent, and responsive to context~\cite{avgar2013environmental}. In particular, a quadruped should be able to interpret the environment and human intent, generate appropriate motion sequences under diverse tasks and interaction scenarios, and support intuitive human–robot interaction through high-level interfaces such as natural language. Despite substantial progress in reinforcement learning~\cite{yue2025rl} and learning-based motion controllers~\cite{zhang2024wococo}, current research faces a fundamental bottleneck: the lack of foundational motion resources that can simultaneously support high-quality motion learning, language–motion semantic alignment, and deployable control, making it difficult to jointly achieve motion diversity, physical realism, and semantic controllability.

Existing approaches either rely on manually engineered gaits and behavior state machines that do not scale to large, long-horizon, multi-style repertoires, or learn from limited motion capture (Mocap) data~\cite{t2qrm}. In the quadruped setting, public datasets are especially small and concentrated (mostly basic primitives such as walk, trot, jump), with scarce interaction behaviors and affective expressions, which restricts open-vocabulary command following and human-facing expressiveness. Moreover, many language-driven methods~\cite{motiongpt3,motionmamba} decouple high-level motion generation from low-level control, producing kinematically plausible but dynamically infeasible motions that fail on hardware.
\begin{figure}[t]
  \centering
  \includegraphics[width=\linewidth]{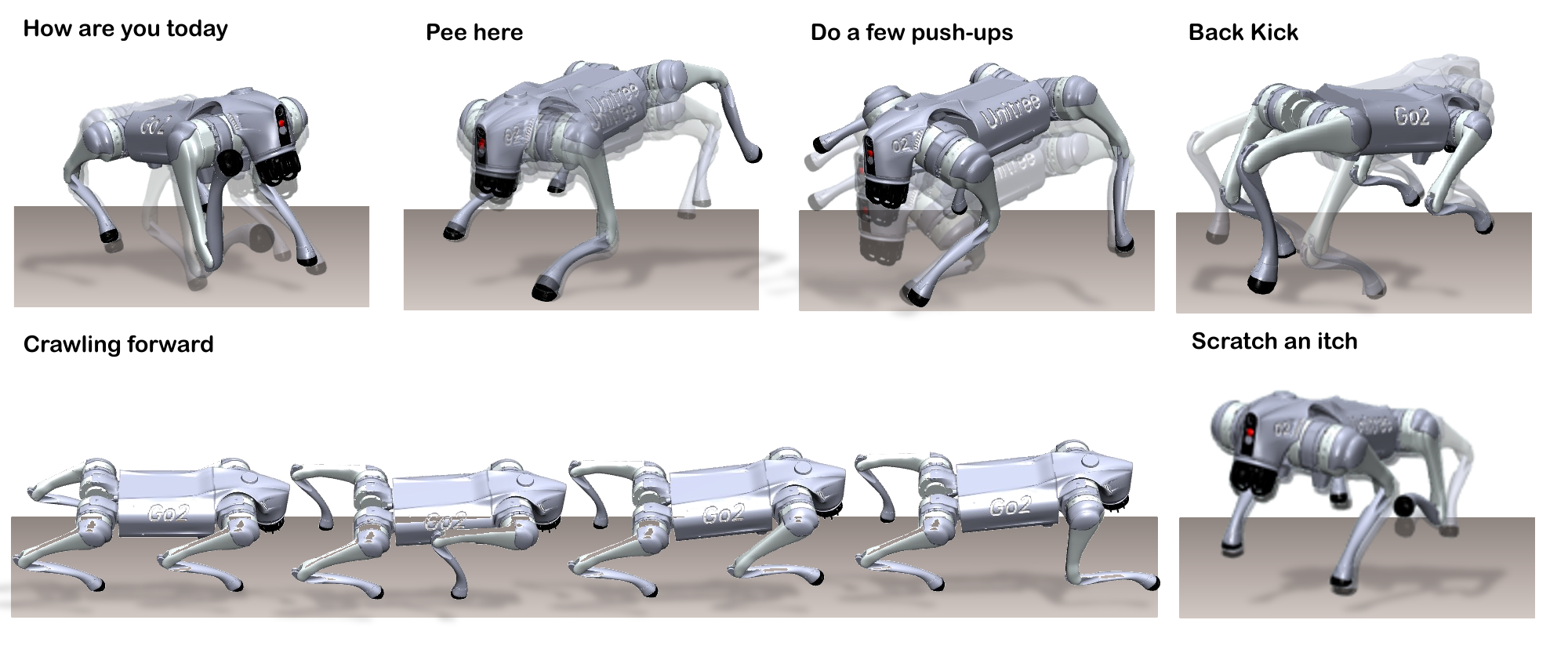}
  \caption{Some motion sequences and representative text annotations sampled from QuadFM. QuadFM exhibit substantial motion diversity. Moreover, the accompanying text annotations are highly interactive and action-oriented. This combination of rich motion coverage and interactive language supervision enables learning controllers that better align physical behaviors with natural, instructive descriptions.
  }
  \vspace{-10pt}
  \label{fig:teaser}
\end{figure}
To address these issues, we introduce Quadruped Foundational Motion (QuadFM), a large-scale, ultra-high-fidelity motion-language dataset designed to unify foundational skills, expressive behaviors, and language grounding. QuadFM is built via a multi-source pipeline (Mocap, artist design, video generation, teleoperation), yielding a broad distribution over gaits, athletic maneuvers, interactive behaviors, and affective motions. It contains 11,784 curated motion clips, each with three layers of text: (i) fine-grained action labels, (ii) context-aware interaction narratives, and (iii) executable natural-language commands, totaling 35,352 descriptions. We further validate its physical fidelity and control utility via downstream motion tracking tasks, where policies trained with QuadFM improve stability, tracking performance under identical setups.

Building on QuadFM, we propose Generation-to-Control Reinforcement Learning (Gen2Control RL), a unified framework that jointly optimizes a general motion controller and a text-conditioned motion generator. By incorporating executability constraints and closed-loop tracking objectives during training, Gen2Control couples semantic generation with dynamical consistency and mitigates common sim-to-real failures of decoupled pipelines. The resulting system supports real-time, on-device inference on Unitree Go2 X quadrupedal robot with an NVIDIA Orin, achieving end-to-end latency below $500$\,ms for responsive language-guided interaction. Extensive simulation and real-world experiments demonstrate improved motion diversity and semantic controllability while maintaining stable execution on physical robots.

Our contributions are threefold:
\begin{itemize}
    \item QuadFM Dataset: We propose a large-scale, ultra-high-fidelity motion-language dataset with diverse interactive behaviors (11,784 clips) and triple-layer annotations (35,352 descriptions) for practical quadruped language grounding.
    \item Gen2Control RL Framework: We propose a co-optimization paradigm that jointly trains text-to-motion generation and general motion controller to ensure semantic controllability with dynamical executability.
    \item Deployable Validation: We conduct comprehensive experiments to verify QuadFM’s high quality and diversity through downstream locomotion control and motion generation tasks, and demonstrate real-time, on-device language-driven motion generation and execution on NVIDIA Orin with end-to-end latency below 500 ms.
\end{itemize}

\section{RELATED WORK}
\subsection{Robot Motion Datasets}
Quadrupedal motion datasets play a pivotal role in advancing learning-based motion generation and robust locomotion control~\cite{peng2020learning, hwangbo2019learning}. To circumvent the prohibitive costs associated with real-world data acquisition, mainstream research predominantly employs motion retargeting techniques~\cite{kang2025learning, peng2018deepmimic} to transfer animal motion capture (mocap) data to robotic platforms. Reinforcement learning~\cite{peng2020learning, han2024lifelike} is subsequently leveraged to bridge the embodiment gap, enabling robots to master lifelike, bio-inspired agility.

While the humanoid robotics domain has capitalized on large-scale databases such as AMASS~\cite{mahmood2019amass} and HumanML3D~\cite{guo2022generating} to develop advanced whole-body imitation systems—including OmniH2O~\cite{he2024omnih2o}, TWIST~\cite{ze2025twist}, and UniAct~\cite{uniact}—and achieve complex physical interactions via retargeting frameworks like GMR~\cite{araujo2025retargeting} and OmniRetarget~\cite{yang2025omniretarget}, high-quality 3D motion datasets dedicated to quadrupedal robots remain conspicuously scarce. Existing quadrupedal resources are largely confined to a limited set of stylized motion primitives~\cite{zhang2018mode}, resulting in a narrow distribution of motor skills. This restricted behavioral repertoire struggles to support the complex, text-driven tasks explored by recent frameworks like SayTap~\cite{tang2023saytap} and T2QRM~\cite{t2qrm}. Ultimately, this profound data scarcity severely bottlenecks the imitation learning capabilities and generalization potential of quadrupedal robots in diverse, semantically rich environments.

\subsection{Text Motion Generation}
Recent progress in human motion generation has been largely driven by text-conditioned synthesis, with surveys systematizing the field by architecture (e.g., VAE-\cite{motiongpt3}, diffusion-\cite{diffvl}, and hybrid-based) and by motion representation (discrete vs. continuous). Beyond standard text-to-motion pipelines, newer unified motion–language frameworks (e.g., MotionGPT3\cite{motiongpt3}) aim to preserve strong language intelligence while generating high-fidelity motions, robotics-oriented systems (e.g., UniAct\cite{uniact}) connect multimodal instruction understanding to low-latency whole-body execution, and efficiency-focused designs (e.g., Motion Mamba\cite{motionmamba}, Efficient Motion Diffusion Models\cite{emdm}) target long-sequence generation and fast inference.

In contrast, quadrupedal motion generation remains less explored. T2QRM\cite{t2qrm} proposes a text-driven framework to synthesize realistic, diverse animal-like quadruped motions in simulation, while integrating reinforcement learning policies to improve adaptability for physical deployment. It further introduces a FrameEncoder for variable-length, long-sequence generation under high-frequency control demands, and releases the DogML dataset pairing dog/quadruped motions with action classes and textual descriptions to support motion–language learning in this domain.

Motivated by these gaps, our work focuses on quadrupedal motion generation, aiming to advance text-driven (and more broadly language-conditioned) motion synthesis for four-legged robots beyond the predominantly humanoid-centered literature.

\subsection{Motion Controllers}
The development of whole-body control for humanoids has historically evolved from computer animation, where reinforcement learning serves as a core tool for achieving precise replication of reference movements~\cite{luo2023perpetual, peng2018deepmimic, peng2021amp, serifi2024vmp} . To handle increasingly sophisticated tasks, researchers have employed adversarial latent spaces~\cite{cui2024anyskill, peng2022ase} and specialized policy architectures~\cite{pan2025tokenhsi, tessler2024maskedmimic} that allow for complex behavior generation via modular inputs~\cite{chen2024taming, wang2025skillmimic, xu2023composite}. Modern frameworks have further advanced this by coupling diffusion-based trajectory planning~\cite{tevet2024closd} with resilient low-level controllers~\cite{fu2024humanplus, he2024omnih2o, he2025asap, he2025hover, ji2024exbody2, li2025amo, ze2025twist, zhang2024wococo}. These systems effectively process a wide array of control signals, ranging from directional navigation~\cite{bloesch2022towards, duan2021learning, radosavovic2024real} and end-effector positioning~\cite{he2024omnih2o, li2025clone} to semantic instructions~\cite{shao2025langwbc, xue2025leverb}.

Despite these advances, quadrupedal whole-body coordination remains under-explored compared to bipedal platforms, primarily due to the lack of large-scale datasets for non-locomotory behaviors. We address this resource gap by curating an expansive quadrupedal motion dataset and proposing a unified framework that bridges generative motion synthesis with high-fidelity physical tracking. This pipeline enables quadrupedal robots to transcend basic gaits and perform diverse, coordinated whole-body tasks.

\section{METHOD}

\begin{figure*}       
	\centering
    \includegraphics[scale=0.2]{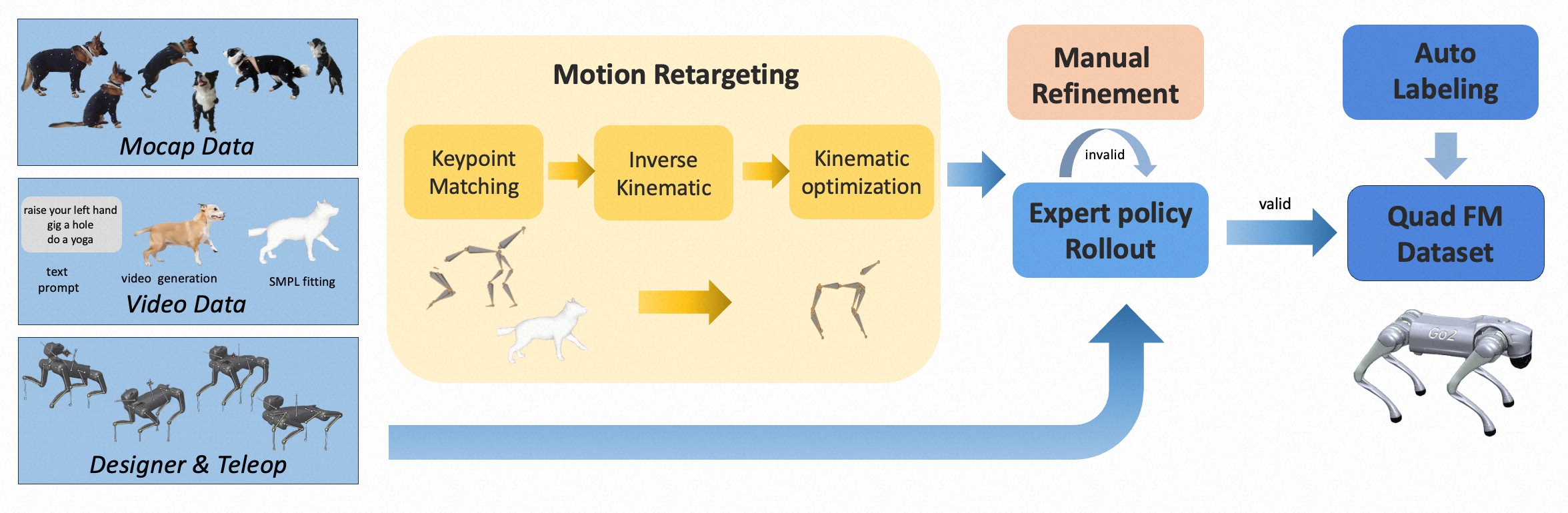}  
	\caption{Data acquisition pipeline for QuadFM dataset}   
	\label{fig:dataset_workflow}
\end{figure*}


\subsection{The QuadFM Dataset}

As illustrated in Fig.~\ref{fig:dataset_workflow}, QuadFM is built through a unified pipeline that starts from motion acquisition, proceeds to robot-oriented motion processing, and ends with dense language grounding. To close the biology-to-robot embodiment gap while supporting rich language-motion learning, we present QuadFM from three perspectives: motion acquisition, robot-executable motion processing, and textual annotation \& dataset statistics.

\subsubsection{Motion Acquisition}
To achieve broad behavioral coverage and high motion fidelity, we adopt a multi-source acquisition strategy:

\paragraph{Motion Capture (Real Dogs)}
As the backbone of the dataset, we capture high-fidelity mocap of real dogs performing fundamental gaits (e.g., walk, trot, pace) and common transitions in controlled studio environments.

\paragraph{Video-to-Motion Generation (Scalable Behaviors)}
To scale behavioral diversity efficiently, we build an automated video-driven synthesis pipeline. We first generate diverse motion video prompts using Qwen2.5VL-72B~\cite{qwen25}, and feed them into Wan~\cite{wan2025wan} to synthesize canine motion videos covering instinctive and social behaviors (e.g., pouncing toward humans, ground scratching, leg lifting, and bipedal begging). We then apply animal pose estimation~\cite{lyu2025animer, zhong20264d} to recover SMAL-based canine poses.

\paragraph{Artist-Designed Motions (Intent-Driven \& Stylized)}
Professional animators craft intent-driven and highly stylized interactive motions (e.g., dancing, joyful bounding) through keyframe animation and interactive authoring based on motion animation principles, complementing behaviors that are rare or difficult to elicit reliably from animals on cue.

\paragraph{Teleoperation Recording (Robot-native Motions)}
We additionally collect robot-native motions via teleoperation: a human operator directly controls the quadruped robot, while the system logs onboard signals such as the robot's joint DOF values and other proprioceptive states. This source provides motions that are naturally executable on hardware and improves coverage of interactive skills.

\subsubsection{Robot-Executable Motion Processing}
Raw animal mocap and video-extracted poses cannot be deployed directly on quadrupedal robots due to substantial morphological and physical mismatches (e.g., bone lengths, joint topology, mass distribution, and actuation limits). We therefore apply an offline kinodynamic workflow to convert raw motions into physically executable robot trajectories.

\paragraph{Kinematic Retargeting}
We map homologous semantic keypoints $\mathcal{K}$ and scale trajectories to the robot morphology ($\hat{p}_{i,t}$). At frame $t$, the robot joint configuration $q_{t}^{*}$ is computed via inverse kinematics:
\begin{equation}
q_{t}^{*} = \arg \min_{q} \sum_{i \in \mathcal{K}} w_{i} \|FK_{i}(q) - \hat{p}_{i,t}\|^{2}
+ w_{reg} \|q - q_{t-1}^{*}\|^{2},
\end{equation}
subject to hardware joint limits. Larger weights $w_i$ on end-effectors emphasize contact fidelity to reduce foot skating, while $w_{reg}$ enforces temporal smoothness.


\paragraph{Physical Motion Correction via RL Tracking}
Purely kinematic retargeting does not guarantee dynamic feasibility. To enforce rigid-body dynamics and realistic contact behavior, we train an RL imitation policy~\cite{he2025asap} to track the retargeted trajectories in a physics simulator:
\begin{equation}
r_{track} = \exp\left(-\sum_{j} c_{j} \|x_{j} - \hat{x}_{j}\|^{2}\right).
\end{equation}
Closed-loop execution introduces physical feedback, producing stable contact dynamics and gravity-consistent motions. We automatically prune rollouts with severe slippage or inconsistency, and experts further refine high-value samples to form a high-precision motion library.

\subsubsection{Textual Annotation and Dataset Summary}
To enable rich language-motion grounding, each curated clip is paired with triple-layer textual annotations:
(1) fine-grained action labels, such as \textit{``Raise your right hand to head height and swing it back and forth three times.''}; 
(2) contextual interaction scenarios, such as \textit{``How are you?''};
(3) executable natural-language commands, such as \textit{``Raise your hand."}.
In total, QuadFM contains over 35,352 semantically rich descriptions.


\subsection{Gen2Control RL Framework}

\subsubsection{Problem setup} At each time step the robot observes proprioceptive state \(s_t\in\mathcal{S}\) and receives a text instruction \(c\). The system outputs a horizon-\(H\) motion sequence \(\tau=\{x_{t:t+H}\}\)  and the low-level controller outputs torques/actions \(u_t\in\mathcal{U}\).

\subsubsection{Imitation pretraining}
We initialize the motion generator with MotionGPT3~\cite{motiongpt3}. A motion VAE encodes a motion clip $m_{1:M}$ into a latent $z=E(m_{1:M})$ and reconstructs it by $D(z)$. Conditioned on text tokens and current pose features, the generator learns a diffusion model $p_\theta(z\mid c,s_t)$ and decodes a reference motion:
\begin{equation}
z \sim p_\theta(z\mid c,s_t),\qquad \hat m_{1:M}=D(z)
\end{equation}
trained on QuadFM using the standard diffusion loss $\mathcal{L}_{\text{diff}}$.

\subsubsection{Gen2Control RL Training}
Pure generation does not guarantee physical executability. We therefore \emph{jointly} train: (i) a language-conditioned motion generator, and (ii) a general motion tracking policy that executes generated motions robustly.

\paragraph{Reference motion sampling.}
The motion generator defines a distribution over reference motions via the latent diffusion model. We sample a latent $z^{(i)}\sim p_\theta(z\mid c,s_t)$ and decode it to a reference motion $\hat m_{1:M}=D(z)$.

\paragraph{Motion tracking policy}
Given the current proprioception and a short future window of the reference motion, the tracking policy outputs low-level actions:
\begin{equation}
u_k=\pi_\phi\!\left(s_k,\tau^{(i)}\right),\qquad s_{k+1}=f(s_k,u_k).
\end{equation}
We train $\pi_\phi$ with PPO using a tracking-style reward (tracking error, energy/smoothness, stability, and termination penalties). We follow ASAP-style training~\cite{he2025asap} with (1) asymmetric actor-critic (critic uses privileged simulation states; actor uses onboard proprioception only), and (2) domain randomization over key dynamics parameters to enable sim-to-real transfer.

\paragraph{Executability feedback to the generator}
The rollout return $R^{(i)}$ (high when the policy tracks well without slipping/falling) is used to bias the generator toward motions that are both instruction-consistent and trackable. Concretely, we apply a policy-gradient style update on the generator using $\log p_\theta(z^{(i)}\!\mid c,s_t)$ weighted by advantage, as detailed in Alg.~\ref{alg:ppo_text2motion}. The overall training alternates between PPO updates for $\pi_\phi$ and periodic RL updates for the generator, while retaining diffusion supervision to preserve semantic diversity.

\begin{algorithm}[t]
\caption{Text2Motion-enhanced PPO (joint training)}
\label{alg:ppo_text2motion}
\begin{algorithmic}[1]
\Require Env $\mathcal{E}$; policy $\pi_\theta$; value $V_\phi$; generator $\mathcal{R}_\tau$
\Require Iterations $N_{\text{iter}}$, rollout length $N_{\text{steps}}$, PPO epochs $N_{\text{epochs}}$
\Require PPO/GAE hyperparams $\Omega$; generator refresh/update period $K_{\mathcal{R}}$
\Ensure Updated $\theta,\phi,\tau$
\State Init optimizers $\mathcal{O}_\pi,\mathcal{O}_V,\mathcal{O}_{\mathcal{R}}$; buffer $\mathcal{D}$
\State $\mathbf{s}\gets \mathcal{E}.\mathrm{reset}()$ \Comment{state $\mathbf{s}$}
\State $G\gets 0$; $\mathcal{G}\gets[\,]$ \Comment{episodic return $G$, list $\mathcal{G}$}
\For{$i\gets 1$ \textbf{to} $N_{\text{iter}}$}
  \If{$i \equiv 0\ (\mathrm{mod}\ K_{\mathcal{R}})$} 
    \State $\mathbf{m}\gets \mathcal{R}_\tau(\text{text})$ \Comment{motion set $\mathbf{m}$}
    \State $\mathbf{s}\gets \mathcal{E}.\mathrm{reset}()$; $G\gets 0$; $\mathcal{G}\gets[\,]$
  \EndIf
  \For{$k\gets 1$ \textbf{to} $N_{\text{steps}}$} \Comment{on-policy data collection}
    \State $\mathbf{s}^a,\mathbf{s}^c\gets \mathrm{split}(\mathbf{s})$ \Comment{actor/critic inputs}
    \State $\mathbf{a}_k,\log p_k\gets \pi_\theta(\mathbf{s}^a)$ \Comment{action, log-prob}
    \State $v_k\gets V_\phi(\mathbf{s}^c)$ \Comment{value estimate}
    \State $(\mathbf{s}',r_k)\gets \mathcal{E}.\mathrm{step}(\mathbf{a}_k)$
    \Comment{reward $r_k$}
    \State Store $(\mathbf{s}^a,\mathbf{s}^c,\mathbf{a}_k,r_k,d_k,v_k,\log p_k)$ in $\mathcal{D}$
    \State $G\gets G+r_k$; \ append $G$ to $\mathcal{G}$
    \State $\mathbf{s}\gets \mathbf{s}'$
  \EndFor
  \State \textbf{PPOUpdate}$(\pi_\theta,V_\phi,\mathcal{D};N_{\text{epochs}},\Omega)$ 
  \If{$i \equiv K_{\mathcal{R}}-1\ (\mathrm{mod}\ K_{\mathcal{R}})$} \Comment{generator RL step}
    \State $b\gets \mathrm{EMA}(\mathcal{G})$ \Comment{baseline $b$}
    \State $A_{\mathcal{R}}\gets \mathcal{G}-b$ \Comment{generator advantage}
    \State $L_{\mathcal{RL}}\gets -\mathbb{E}\!\left[\log p_\tau(\mathbf{m})\,A_{\mathcal{R}}\right]$
    \Comment{RL loss}

    \State $L_{\mathcal{R}} \gets L_{\text{recon}} + L_{\text{RL}}$ 
    \State Step $\mathcal{O}_{\mathcal{R}}$ on $L_{\mathcal{R}}$
  \EndIf
  \State Clear $\mathcal{D}$
\EndFor
\end{algorithmic}
\end{algorithm}

\section{Experiments}

\subsection{Dataset Study} 
\begin{table*}[htbp]
\centering
\caption{Detailed Statistical Breakdown of the QuadFM Dataset vs. Baseline}
\resizebox{\textwidth}{!}{
\label{tab:dataset_breakdown}
\begin{tabular}{ccccccc}
\toprule
\multirow{2}{*}{\textbf{Motion Category}} & \multirow{2}{*}{\textbf{Data Source}} & \multirow{2}{*}{\textbf{Semantic Subset / Behavior Types}} & \multicolumn{2}{c}{\textbf{QuadFM (Ours)}} & \multicolumn{2}{c}{\textbf{DogML}} \\
\cmidrule(lr){4-5} \cmidrule(lr){6-7}
 &  &  & \begin{minipage}{2cm}\centering\textbf{Clips} \end{minipage}& \begin{minipage}{2cm}\centering \textbf{Duration}\end{minipage} & \begin{minipage}{2cm} \centering \textbf{Clips} \end{minipage}&  \begin{minipage}{2cm}\centering\textbf{Duration}\end{minipage} \\
\midrule
\multirow{2}{*}{\textbf{Locomotion}} 
 & Motion Capture & Basic Gaits (Walking, Trotting, etc.) & 7998 & 10.02 h & 4,024$^*$ & 11.03 h \\
 & Video Generation & Diverse In-the-wild Behaviors & 1,392 & 1.62 h & 0 & 0.00 h \\
\midrule
\multirow{2}{*}{\begin{tabular}[c]{@{}l@{}}\textbf{Interaction} \\ \textbf{\& Emotion}\end{tabular}} 
 & Teleoperation & Natural (Greeting), Sad (Cautious) & 696 & 4.74 h & 0 & 0.00 h \\
 & Artist Design & Happy (Dancing, Excited), etc. & 1,698 & 3.89 h & 0 & 0.00 h \\
\midrule
\textbf{Total} & \textbf{--} & \textbf{All Sources / Comprehensive} & \textbf{11,784} & \textbf{20.27 h} & \textbf{4,024$^*$} & \textbf{11.03 h} \\
\bottomrule
\multicolumn{7}{l}{\rule{0pt}{3ex}\footnotesize $^*$ \textit{Note:} DogML\cite{t2qrm} count includes redundant retargeted sequences. For a consistent comparison of behavioral diversity, we report the \textbf{4,024} unique motion events.}
\end{tabular}%
}
\vspace{-1em}
\end{table*}
To demonstrate the comprehensiveness and physical fidelity of QuadFM, we conduct a systematic comparison with existing quadruped datasets, focusing on DogML~\cite{t2qrm}. Our evaluation targets micro-scale kinematic diversity. Directly comparing raw frame counts can be misleading, since long recordings of repetitive gait cycles (e.g., extended pacing) inflate the apparent scale without adding new behaviors. To obtain a fair measure of diversity, we apply a kinematic deduplication procedure: we segment continuous sequences and compute Dynamic Time Warping (DTW) distances in joint-velocity space, then merge highly similar cyclic motions. This yields a normalized metric, the number of Unique Atomic Motions, which we use throughout the comparison.

To further visualize structural diversity, we extract for each clip a dynamic feature vector $\mathbf{f}\in\mathbb{R}^{D}$ consisting of root velocities, 12-DoF joint positions, and foot-contact states, and project these features to 2D using UMAP.

\paragraph{Scale and hierarchical composition} The breakdown in Table~\ref{tab:dataset_breakdown} shows that QuadFM contains 11,784 unique atomic motions and 3.64M frames (normalized to 50\,Hz), substantially exceeding DogML. More importantly, QuadFM is organized with a hierarchical taxonomy. DogML is heavily concentrated in locomotion, dominated by steady-state periodic gaits. In contrast, while QuadFM also provides a strong locomotion foundation, it additionally includes a large Interaction \& Emotion subset (e.g., greeting, cautious/sad pacing, excited dancing and bounding), which is entirely absent in DogML (zero clips).

\begin{figure}[htbp]
    \centering
    \includegraphics[width=0.5\textwidth]{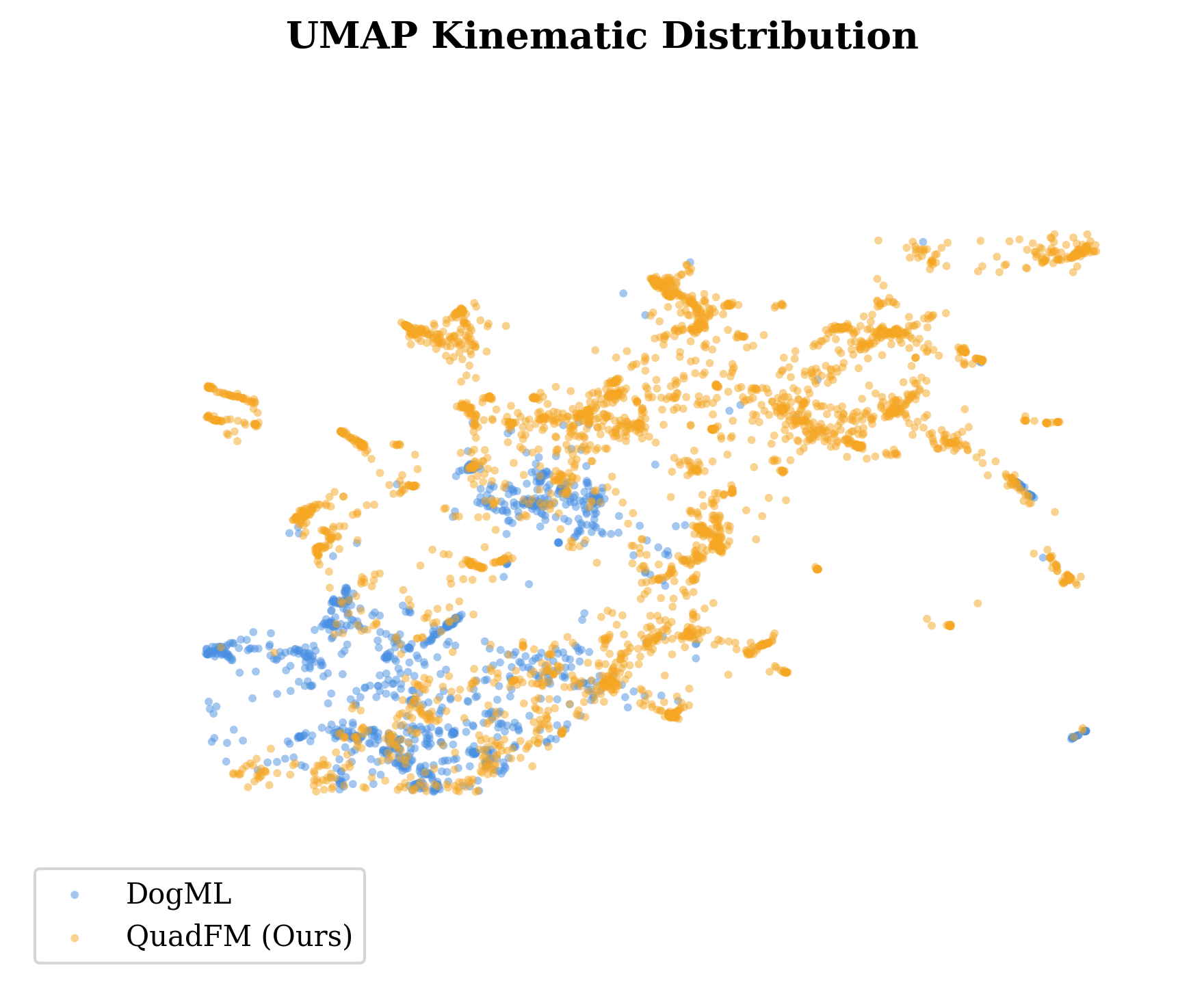}
    \caption{Comprehensive physical analysis of the datasets. UMAP projection of kinematic features demonstrates that QuadFM expansively covers novel dynamic and expressive motion manifolds, effectively preventing the localized mode collapse observed in DogML.}
    \label{fig:dataset_analysis}
\end{figure}

\paragraph{Kinematic distribution and physical diversity} The UMAP embedding in Fig.~\ref{fig:dataset_analysis} validates that QuadFM provides substantially broader continuous kinematic coverage. DogML forms compact clusters corresponding mainly to standardized gait patterns, indicating limited variability despite its duration. QuadFM not only overlaps these baseline clusters but also expands into distinct peripheral regions, which correspond to highly dynamic and expressive behaviors (e.g., jumping, stretching, and dancing). This expanded manifold indicates richer state coverage for downstream motion generation and control.

\begin{figure*}[t]
    \centering
    \includegraphics[width=\textwidth]{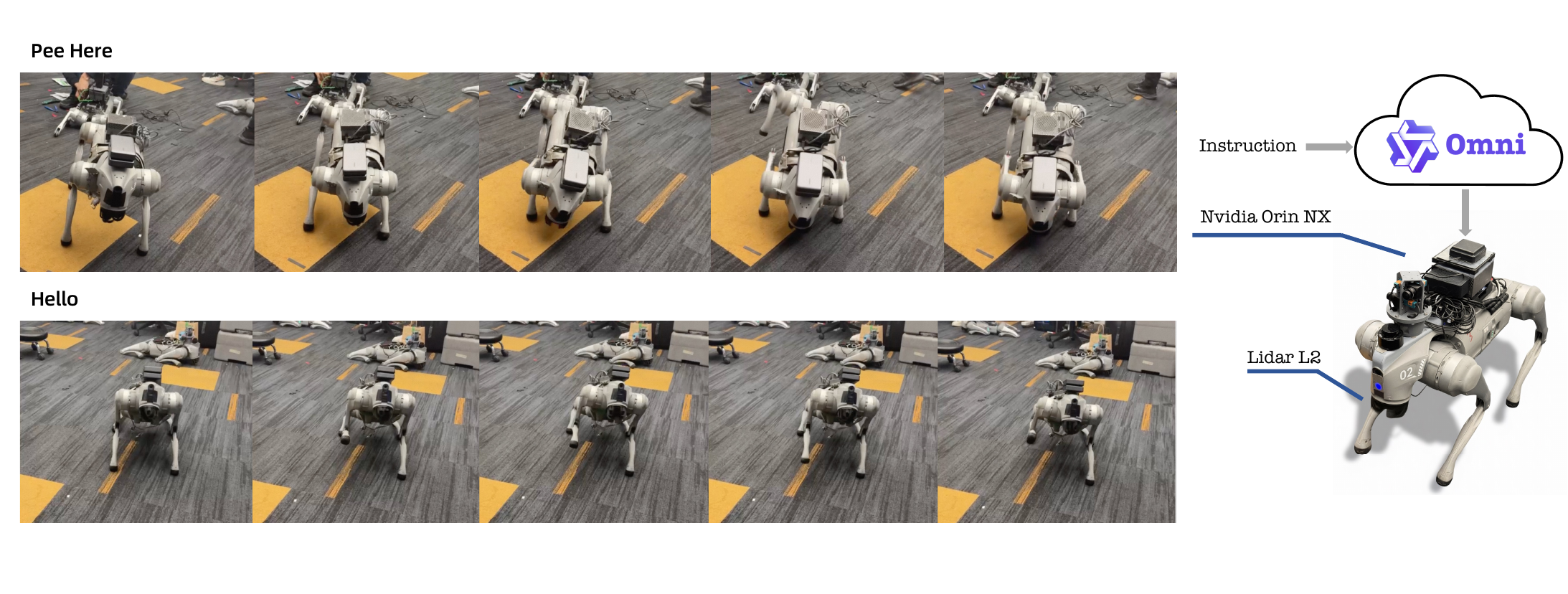}
    \caption{Real-time motion generation on NVIDIA Jetson Orin.}
    \label{fig:realworld}
\end{figure*}

\subsection{Generation-to-Control Reinforcement Learning}
\subsubsection{Implementation Details}
In Stage I (imitation learning), we train the model using the MotionGPT3~\cite{motiongpt3} framework on a single AMD MI308X GPU with a batch size of 24, for approximately 48 hours. The imitation-learning pipeline consists of two steps. (1) VAE pretraining: we pretrain the VAE using AdamW~\cite{kingma2014adam} with a learning rate of \(2\times10^{-4}\) for 300 epochs. (2) Language model training: we freeze the VAE and train the LM using AdamW with a learning rate of \(1\times10^{-4}\) for 300 epochs.

In Stage II (Gen2Control joint training), we perform RL-based joint post-training on two NVIDIA RTX 3090 GPUs, with both the VAE and LM unfrozen. The motion generator learning rate is \(1\times10^{-4}\), while the actor and critic learning rates are \(1\times10^{-3}\). Training runs for 20,000 iterations. We use 4,096 parallel environments with a rollout length of 24 steps. Both the actor and critic are 4-layer MLPs with hidden dimensions 2048, 1024, 512, and 256. During joint training, for each action sequence sampled from motion generator, we update the generic locomotion controller for 100 optimization steps, and use the resulting rollout return/tracking loss to optimize motion generator.
\subsubsection{Evaluation Metrics}
Following prior work~\cite{cui2025grove,cui2024anyskill}, we adopt human evaluation as our primary assessment protocol. We selected 10 open-vocabulary text descriptions that require comprehensive whole-body motions and are not covered by the Gen2Control RL training data. To evaluate the generated motions, we recruited 30 participants to rate each sample on text–motion alignment, smoothness, naturalness, and stability using a 0–9 scale.
\subsubsection{Real World Deployment}
Our end-to-end real-robot deployment runs on a Unitree GO2 X quadruped, where both the motion generator and the low-level controller are deployed on an NVIDIA Jetson Orin, as shown in ~\ref{fig:realworld}. The motion generator runs at 2Hz to produce motion commands, while the generic locomotion controller executes them at 50Hz for stable tracking. For voice-driven interaction, ASR is performed in the cloud using Qwen-Omni~\cite{omni}, and the recognized instruction is sent to the robot over the network; the overall latency from speech input to motion execution is approximately 0.5s.

\subsubsection{Experimental Results}
We compare our method against MotionGPT3 and a closely matched ground-truth motion reference. As reported in Tab. ~\ref{tab:gen2control}, Gen2Control consistently outperforms MotionGPT3 across all metrics, producing motions that are notably closer to the ground truth. Note that the “ground-truth” (GT) reference in our study is selected by retrieving, from the training set, the text description with the highest similarity to each test instruction and then using its paired motion as the reference; therefore, the GT does not necessarily perfectly match the test text and its text–motion alignment score may be lower than the maximum. In addition, all motions—including the GT reference—are executed and rated on the Unitree Go2 platform, and the human scores are influenced by subjectivity as well as by the general motion controller’s tracking performance and real-world environmental factors. As a result, the GT motions may also receive non-saturated ratings in naturalness, smoothness, and stability, rather than achieving full scores.

\begin{table}[htbp]
\centering
\caption{Quantitative evaluation of Gen2Control RL.}
\label{tab:gen2control}
\resizebox{\columnwidth}{!}{%
\begin{tabular}{ccccc}
\toprule
               & Stability$\uparrow$ & Text-Motion Alignment$\uparrow$ & Smoothness$\uparrow$ & Naturalness$\uparrow$ \\ \midrule
MotionGPT3~\cite{motiongpt3}     &6.27   &  5.77   &  5.62 & 5.54 \\
Gen2Control RL(ours) & 7.58 &     7.98   &   7.46  &     7.40    \\\midrule
Related GT     &    8.01   &   8.21     &      7.82     &        7.69     \\ \bottomrule
\end{tabular}
}
\end{table}

\subsection{Ablation Study}
To analyze how dataset configuration affects performance, we perform ablation studies focusing solely on the control policy within IssacSim simulator. The validation set was formed by randomly reserving a 1,888.29-second subset. Tracking performance is quantitatively evaluated using the following two metrics:

\begin{itemize}
\item Mean Joint Position Error (MJPE). This metric measures the pose tracking fidelity at the joint level. It is defined as the Mean Absolute Error (MAE) between the reference joint angles and the simulated robot's joint angles, averaged over all actuated joints ($J$) and all time steps ($T$) of the motion sequence.


\item Mean Body Position Error (MBPE). This metric evaluates the global tracking accuracy of the robot's physical structure. It is computed as the Mean Absolute Error (MAE) of the global Cartesian coordinates $(x, y, z)$ for all rigid bodies $B$ relative to the reference motion. The error is averaged over all rigid bodies, coordinate axes, and time steps.


\end{itemize}

\begin{table}[htbp]
\centering
\caption{Ablation study on different dataset configuration. We report MJPE (rad) and MBPE (m) on both our validation set and the DogML~\cite{t2qrm} validation set. ($\downarrow$) indicates lower values are better. Best results are highlighted in \textbf{bold}.}
\label{tab:data_scaling}
\resizebox{\columnwidth}{!}{%
\begin{tabular}{l cccc}
\toprule
\multirow{2}{*}{\textbf{Training Data}} & \multicolumn{2}{c}{\textbf{Ours Val}} & \multicolumn{2}{c}{\textbf{DogML~\cite{t2qrm} Val}} \\
\cmidrule(lr){2-3} \cmidrule(lr){4-5}
 & \textbf{MJPE} ($\downarrow$) & \textbf{MBPE} ($\downarrow$) & \textbf{MJPE} ($\downarrow$) & \textbf{MBPE} ($\downarrow$) \\
\midrule
DogML~\cite{t2qrm} (100\%) & 0.0964 & 0.2832 & 0.1331 & 0.1009 \\
\midrule
Ours (only teleoperation) & 0.0884 & 0.0806 & 0.2571 & 0.1304 \\
Ours (only artist design) & 0.0871 & 0.0770 & 0.1640 & 0.1072 \\
Ours (only motion capture) & 0.0968 & 0.2338 & 0.2331 & 0.1201 \\
Ours (only video generation) & 0.1507 & 0.1306 & 0.2368 & 0.1149 \\
Ours & \textbf{0.0712} & \textbf{0.0744} & 0.1372 & 0.1017 \\
\midrule
Ours + DogML~\cite{t2qrm} (25\%) & 0.0741 & 0.0794 & 0.1224 & 0.0945 \\
Ours + DogML~\cite{t2qrm} (50\%) & 0.0776 & 0.0854 & 0.1203 & 0.0937 \\
Ours + DogML~\cite{t2qrm} (100\%)& 0.0799 & 0.0895 & \textbf{0.1197} & \textbf{0.0916} \\
\bottomrule
\end{tabular}
}
\end{table}

Table~\ref{tab:data_scaling} reports results under three training configurations: (1) training only on DogML~\cite{t2qrm}, (2) assessing the individual and combined effects of different data sources, and (3) co-training our dataset with varying proportions of DogML.

As shown in Table~\ref{tab:data_scaling}, training solely on DogML generalizes poorly to our validation set, yielding a notably high MBPE. In contrast, training on our data achieves comparable performance on the DogML validation set to the DogML-trained model, indicating the high quality and strong generalization of our dataset. Experiments on individual data components show that each source contributes useful information, while combining all sources gives the best overall performance, confirming their complementarity.

With co-training (bottom of Table~\ref{tab:data_scaling}), we observe clear synergy. We apply proportional random sampling on DogML. Adding only 25\% DogML to our dataset already surpasses training on 100\% DogML. Performance on the DogML validation set peaks when using 100\% of both datasets. Meanwhile, increasing the DogML proportion slightly degrades performance on our validation set, which can be attributed to domain distraction or distribution shift introduced by DogML data. Overall, co-training further demonstrates the diversity and quality of our data.

\section{Conclusion}

This paper introduces QuadFM, a large-scale, ultra-high-fidelity motion–language dataset for quadrupedal robots designed to unify foundational locomotion skills,       expressive/interactive behaviors, and language–motion semantic alignment. Through a multi-source data construction and rigorous curation pipeline, together with triple-layer textual annotations (action/context/command), QuadFM substantially increases both motion diversity and semantic density, enabling learning and evaluation of open-vocabulary, human-facing quadruped behaviors. Building on this dataset, we further present a joint co-optimization framework Gen2Control that trains a text-conditioned motion generator together with a general motion controller under closed-loop executability objectives, mitigating the common instability and real-robot failure modes of decoupled generate-then-track pipelines. Extensive simulation and hardware experiments validate that QuadFM improves stability, tracking accuracy, and robustness for locomotion policy learning, while supporting real-time on-device language-driven motion generation and execution with sub-500 ms end-to-end latency. 
\section{Future Work}
Looking forward, we plan to expand QuadFM toward richer compositional semantics and more complex interaction scenarios (e.g., multi-intent commands, temporal constraints, and multi-agent behaviors). We also aim to advance closed-loop language grounding via interactive prompting, clarification, and online correction with human feedback, and to establish standardized evaluation protocols for motion–language alignment, expressiveness, physical executability, and real-robot reliability, complemented by user studies on perceived naturalness and intent interpretability.





\bibliographystyle{IEEEtranS}
\bibliography{IEEEfull}

\end{document}